\documentclass[11pt,letterpaper]{article}
\usepackage[letterpaper, total={6.5in, 9.5in}]{geometry}

\usepackage{epsfig}
\usepackage{amsmath}
\usepackage{amssymb}
\usepackage{amsfonts}
\usepackage{color}
\usepackage{url}
\usepackage{graphicx}
\usepackage{subfig}
\usepackage{enumitem}
\usepackage{multirow}
\usepackage{booktabs}
\usepackage[table]{xcolor}
\usepackage{nicefrac}
\usepackage{microtype}
\usepackage{hyperref}
\usepackage{makecell}
\usepackage{environ}
\usepackage[sort]{cite}
\newcommand{\acksection}{\section*{Acknowledgments and Disclosure of Funding}}
\NewEnviron{ack}{%
  \acksection
  \BODY
}

\definecolor{redMod}{HTML}{B45A4A}
\definecolor{redStr}{HTML}{7A1F1F}

\newlength{\figurewidth}
\newlength{\smallfigurewidth}

\begin{document}

\title
{\large
\textbf{Neural Compression for Scientific Particle Data: Perspectives on How Learning Can Help}
}

\author{%
Tian Qiu$^{1}$, Carolina Cuesta-Lazaro$^{2,3}$, Sheng Di$^{4}$ and Haewon Jeong$^{1,3 }$\\[0.5em]
{\small\begin{minipage}{\linewidth}\begin{center}
\begin{tabular}{c}
$^{1}$University of California, Santa Barbara $^{2}$New York University\\
$^{3}$Flatiron Institute $^{4}$Argonne National Laboratory\\
\url{{tian_qiu,haewon}@ucsb.edu}, \url{cc9701@nyu.edu}, \url{sdi1@anl.gov}\\
\end{tabular}
\end{center}\end{minipage}}
}

\maketitle
\thispagestyle{empty}

\begin{abstract}
As particle-based simulations grow in scale, efficient compression of scientific data becomes critical. Neural point cloud compression (PCC) methods excel on computer vision (CV) benchmarks, but do they transfer to scientific particle data? We highlight that scientific particles differ from CV point clouds in fractal dimension: CV point clouds are locally planar, but particles are not. This key difference limits the effectiveness of the octree data structure that underpins most PCC methods. Neural codecs achieve only marginal bitrate reductions over the traditional G-PCC baseline on scientific data, revealing a significant performance gap. Supported by an evaluation spanning domains of cosmology, molecular dynamics, and plasma physics, we outline where the structure of scientific data limits current neural approaches and where learning can help in the future. 
\end{abstract}

\section{Introduction}
Particle-based simulation has become a fundamental tool in the physical sciences. It enables us to build high-fidelity digital twins of complex systems. The same fidelity, however, generates rapidly growing data volumes, making efficient compression an urgent need. Cosmological $N$-body simulations~\cite{Quijote, maksimova2021abacussummit, garrison2021abacussummit} now model the interactions of billions of dark matter particles to study the formation and evolution of large-scale structures in the Universe. Molecular dynamics (MD) simulations~\cite{abraham2015gromacs,karplus2002molecular,hollingsworth2018molecular} resolve atomic-scale phenomena, such as metal deposition~\cite{dai2025molecular}, protein folding~\cite{lindorff2011fast}, and drug interactions~\cite{de2016role}, generating long trajectories of millions of atoms. As computational capabilities and simulation fidelity continue to scale, the resulting datasets routinely reach terabyte scale~\cite{cappello2025lossy}, imposing substantial burdens on storage, data movement, and downstream analysis. Consequently, efficient compression of scientific particle data has become a pressing and shared need across disciplines.

Compression of scientific data has received growing attention in the high-performance computing community, with developments in both structured grid data and unstructured particle data. Compressors for structured grids~\cite{liang2022sz3, lindstrom2014fixed} leverage strong local correlation between storage-adjacent elements in various ways to achieve compression. However, particle data poses a unique challenge: spatial location is part of the data itself, and storage order often breaks spatial correlation. Current methods for particle data~\cite{zhang2025lcp} typically operate on individual coordinates or local neighborhoods, and do not exploit the large-scale spatial structure present in the data, leaving significant room for improvement. 

Recent advances in neural point cloud compression (PCC) in the computer vision community raise a compelling opportunity to significantly improve compression for scientific particle data. Traditional PCC methods are primarily designed for computer vision workloads such as LiDAR scans and human body captures. The dominant standard, MPEG G-PCC~\cite{gpcc2023}, represents point clouds using an octree structure and relies on hand-crafted context models to entropy-code occupancy information. In contrast, neural PCC approaches~\cite{gao2025deep} learn data-driven predictive models, consistently achieving substantially higher compression ratios than G-PCC on standard computer vision benchmarks. These advances prompt a natural question: \textit{can similar gains be realized for scientific particle datasets?} Comparable gains on scientific particle data would dramatically reduce storage and data movement costs, making large-scale particle simulations far more accessible for downstream analysis.

In this paper, we investigate the central question: \textit{\textbf{Do neural compression methods achieve comparable effectiveness on scientific particle data as they do on computer vision point clouds?}} To answer it, we decompose it into three sub-questions:
\begin{enumerate}[nosep, leftmargin=*]
    \item \textit{How does scientific particle data differ from computer vision point clouds?}
    \item \textit{Is the octree an efficient representation for scientific particle data?}
    \item \textit{Given the octree representation, do neural models improve over simple baselines?}
\end{enumerate}
To answer these questions, we prepare publicly available scientific particle datasets for octree-based compression. Subsequently, guided by the sub-questions above, we provide a systematic analysis of where and why neural compression methods succeed or fail on particle data. Together, these contributions offer both practical benchmarking guidelines and actionable insights for the community at the intersection of neural compression and scientific computing.
\begin{figure}[t]
    \centering
    \includegraphics[width=\textwidth]{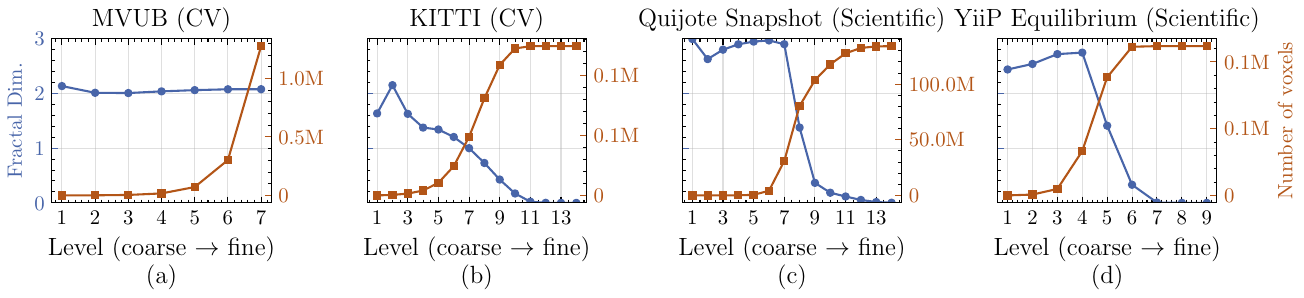}
    \caption{\textbf{Fractal dimension reveals a geometric difference between CV and scientific point clouds.} \textbf{(a) Human body scan.} The dimension remains near 2 at all levels, consistent with locally planar surface geometry. \textbf{(b) Vehicular LiDAR.} Fractal dimension increases from near 0 at fine levels to approximately 2 at coarse levels. This reflects the physical nature~\cite{geiger2013vision} of spinning LiDAR sensors: scan rings appear locally linear at finer levels and appear denser at coarser levels. \textbf{(c) Cosmology $N$-body simulations.} Fractal dimension rises quickly from 0 to 3, indicating a sharp transition from sparse to volume-filling geometry. \textbf{(d) Molecular dynamics simulations} show the same sharp sparse-to-dense transition as cosmology simulations. This contrast illustrates the fundamental geometric difference between CV point clouds and scientific particle data discussed in Sec.~\ref{sec:q1}.}
    \label{fig:fractal}
\end{figure}
\section{Analysis: Are existing neural compressors effective for particle data?}
Our evaluation spans three domains: cosmology, molecular dynamics, and plasma physics. For cosmology, we use the Quijote $N$-body simulations~\cite{Quijote}. For molecular dynamics, we use the YiiP membrane-protein system~\cite{fan2019yiip} simulated with GROMACS. For plasma physics, we use the charged-particle beam dataset generated with WarpX~\cite{WarpX} by Sandberg \textit{et al.}~\cite{sandberg2024synthesizing}. Particle simulations store positions as floating-point numbers, whereas PCC codecs require integer voxel coordinates. We apply uniform scalar quantization $x_i = \lfloor x / \Delta \rceil$, choosing $\Delta$ at or below the physical resolution limit of each domain, so that quantization error is negligible for downstream analyses.
\subsection{How is scientific particle data different from CV point clouds?}\label{sec:q1}
CV point clouds share a common geometric property: from CG models~\cite{chang2015shapenet}, LiDAR scans~\cite{geiger2013vision, pandey2011ford}, to RGB-D captures~\cite{deon20178i, dai2017scannet, mammou2019gpcc}, all CV point clouds describe object surfaces and are locally planar. Scientific particles do not share this property. Particles in scientific simulations represent different physical entities depending on the domain, and their spatial distributions vary greatly. 

We demonstrate this distinction through the lens of \textit{fractal dimension}~\cite{falconer2003fractal}, which characterizes how a point cloud fills space across scales. Consider partitioning the bounding volume into a grid of $2^l\times2^l\times2^l$ cells and marking each cell as occupied if it contains at least one point. Let $N_l$
denote the number of occupied voxels. The fractal dimension,  computed by $d_l = \log_2(N_{l}/N_{l-1})$, is a metric that captures how quickly the number of occupied voxels grows as grid resolution doubles.
While this metric need not be an integer, it reduces to the more familiar Euclidean dimension in regular cases: it evaluates to 3 for solid volumes, 2 for surfaces, and 1 for lines. Due to its coincidence with Euclidean dimension and ease of computation, it has been used as an indicator of point sparsity in the PCC literature~\cite{pang2022grasp, wang2024versatile}. 

Our results are shown in Fig.~\ref{fig:fractal}. CV point clouds representing surfaces have a consistent fractal dimension close to 2. Scientific particle data, by contrast, exhibits different patterns:
\begin{enumerate}[nosep, leftmargin=*]
    \item \textbf{Cosmology $N$-body} simulations~\cite{springel2005cosmological} track how dark matter particles interact under gravity. Over time, particles collapse into a complex web of voids, filaments, sheets, and dense halos~\cite{bond1996filaments}. These structures span a wide range of intrinsic dimensionalities (see Fig.~\ref{fig:fractal}\,(c)).
    \item \textbf{Molecular dynamics} simulations~\cite{abraham2015gromacs, hollingsworth2018molecular} track atom movements in biomolecular systems such as membranes. Atomic positions are governed by interactions at distinct scales: intramolecular forces such as covalent bonds constrain atoms into rigid geometries, while intermolecular forces drive the arrangement of molecules relative to each other (see Fig.~\ref{fig:fractal}\,(d)). 
    \item \textbf{Plasma physics}~\cite{fedeli2022pushing} models charged particles in electromagnetic fields, producing distributions shaped by beam dynamics and instabilities that, again, do not have surface-like structure.
\end{enumerate}

\subsection{Is the octree an efficient representation for scientific particle data?}\label{sec:q2}
The octree is the most common way to represent CV point cloud data for compression~\cite{schnabel2006octree, graziosi2020overview, gao2025deep}. It represents a point cloud by recursively splitting the bounding volume into eight sub-volumes, encoding at each node an 8-bit occupancy code that signals which children are occupied. The problem of compressing plain coordinates \(\{x_i, y_i, z_i\}\) is converted to compressing the breadth-first traversal of the occupancy code sequence \(\{O_k\}\).

The octree representation describes data in a multi-scale hierarchy. This representation offers three advantages: representation-level compression, support for multi-resolution decoding~\cite{knoll2006survey}, and a ladder for context-adaptive entropy coding of occupancy codes. For a point cloud with consistent voxel merging from a fine level to a coarse level, the cost of occupancy codes is easily offset by the rapidly reducing number of octree nodes~\cite{botsch2002efficient, lasserre2019using}. Botsch \textit{et al.}~\cite{botsch2002efficient} showed that for a consistently planar point cloud, the bit-per-point cost of octree is upper-bounded by $8/3 \approx 2.67$. For CV point clouds, these representation-level savings are reflected in the Raw Octree column of Table~\ref{tab:compression_bpp}. 

This advantage diminishes for scientific particle data: $d$ exceeds 2 at coarse levels but rapidly approaches 0. At $d$ close to 0, the octree is no longer efficient, as it achieves data reduction by merging voxels. When no voxels merge, an 8-bit occupancy code signals at most $\log_28=3$ bits of information. Five bits are wasted on each point, and this waste accumulates over multiple levels. 
This waste can be partly recovered by hand-designed flags~\cite{peng2003octree} or learned entropy coding. Still, the inefficiency suggests that octree, despite being predominant for CV, may not suit scientific particle data.

\begin{table*}[t]
\footnotesize
\centering
\caption{\textbf{Neural methods achieve far smaller gains on scientific particle data than on CV data.} Bit-per-point (bpp) by method, with compression ratio in parentheses, computed over post-quantization bitrate. General-purpose scientific compressor SZ3 suffers on particle data, while particle compressor LCP performs on par with neural CV point cloud codecs on scientific data. ``--'' marks entries where the method was not evaluated on the dataset originally.}
\label{tab:compression_bpp}
\begin{tabular}{@{}llccccccc@{}}
\toprule
 & \makecell[l]{Dataset\\(\# test samples)} & \makecell{Post-\\Quant.} & \makecell{Raw\\Octree}
 & G-PCC & RENO & \makecell{Oct-\\Attention} & SZ3 & LCP \\
\midrule
\multirow{5}{*}{\rotatebox{90}{\textit{Scientific}}}
 & Quijote halos (200)     & 54 & 87.40 (0.6) & 35.92 (1.5) & 35.55 (1.5) & 35.78 (1.5) & 60.18 (0.9) & 36.22 (1.5) \\
 & Quijote particles (200) & 54 & 88.37 (0.6) & 36.03 (1.5) & 35.77 (1.5) & 35.92 (1.5) & 59.46 (0.9) & 36.25 (1.5) \\
 & Quijote boxes (1600)    & 45 & 53.89 (0.8) & 23.52 (1.9) & 22.68 (2.0) & 22.82 (2.0) & 40.36 (1.1) & 24.52 (1.8) \\
 & YiiP (900)               & 45 & 59.70 (0.8) & 25.35 (1.8) & 22.89 (2.0) & 23.91 (1.9) & 44.70 (1.0) & 27.08 (1.7) \\
 & WarpX (30)               & 54 & 55.84 (1.0) & 23.74 (2.3) & 23.35 (2.3) & 23.33 (2.3) & 41.25 (1.3) & 24.70 (2.2) \\
\cmidrule(lr){2-9}
\multirow{4}{*}{\rotatebox{90}{\textit{CV}}}
 & MVUB, Phil (245)         & 30 & 2.55 (11.8) & 1.07 (28.0) & --          & 0.79 (38.0) & -- & -- \\
 & 8iVFB, Loot (300)        & 30 & 2.98 (10.1) & 0.95 (31.6) & --          & 0.62 (48.4) & -- & -- \\
 & KITTI (7518)             & 52 & 17.85 (2.9) & 8.95 (5.8)  & 7.32 (7.1)  & --          & -- & -- \\
 & Ford (60)                & 52 & 24.56 (2.1) & 11.72 (4.4) & 9.75 (5.3)  & --          & -- & -- \\
\bottomrule
\end{tabular}
\end{table*}

\subsection{Do neural codecs improve particle data compression?}
To isolate the benefit of learned context models for octree coding, we train two neural compression models on multi-frame datasets curated from the three domains. For Quijote, we derive three variants (halos, particles, and boxes) to cover different geometries. Frames contain roughly 100,000 points to fit GPU memory. For all the codecs, we use one quantization step per dataset, chosen so that quantization error is negligible for downstream analyses. Since our goal is to assess the effectiveness of neural codecs on scientific data in the near-lossless regime where distortion has negligible impact on downstream tasks, a single operating point suffices. We leave rate--distortion analysis of lossy compression to future work. Full curation details are in Appendix~\ref{appendix:data}.

Results in Table~\ref{tab:compression_bpp} show that OctAttention~\cite{fu2022octattention} and RENO~\cite{you2025reno} achieve surprisingly small gains over G-PCC: at best a 10\% bitrate reduction on scientific data (RENO on YiiP), compared with around 17\% on LiDAR and 30\% on human scans.

\begin{figure}[t]
\centering
\subfloat[]{\label{fig:mvub_parent_conditional}%
\includegraphics[width=0.47\columnwidth]{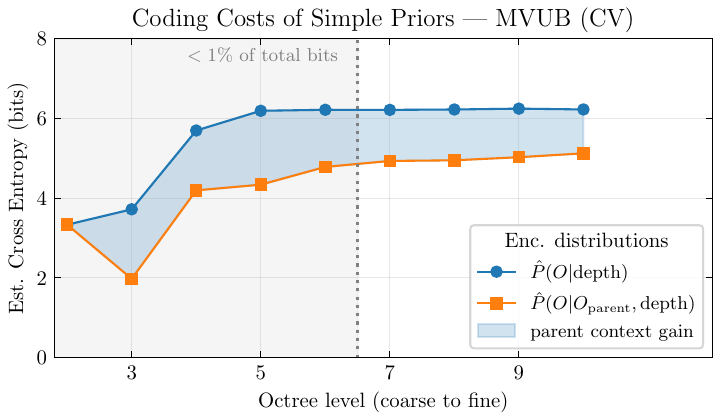}}
\qquad
\subfloat[]{\label{fig:quijote_halos_parent_conditional}%
\includegraphics[width=0.47\columnwidth]{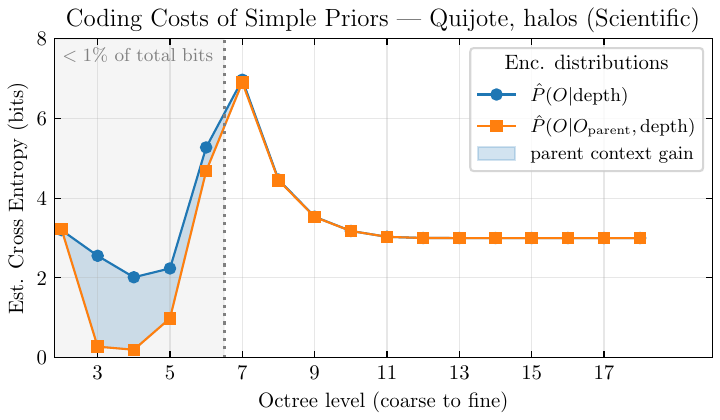}}
\caption{\textbf{CV point cloud compression benefits across levels from conditioning on a single parent node, while particle datasets only benefit from it at coarser levels.} \textbf{(a)} On MVUB (human body scans), points concentrate on surfaces, so a child's occupancy can be strongly predicted by its parent's: conditional entropy drops significantly across levels. \textbf{(b)} Quijote halo data benefits from the same model only on coarser levels. Gains in these levels are less significant, because the number of voxels to code here is negligible compared to the entire octree. Gray area indicates levels that contribute less than 1\% of the total bitstream -- gains in these levels have little effect on overall bpp.}
\label{fig:parent_conditional_comparisons}
\end{figure}

To build an intuition for why neural context models yield only marginal gains, we examine the core problem of compressing the occupancy code sequence $\{O_k\}$ through a simple but informative test: conditioning on the parent node. In Fig.~\ref{fig:parent_conditional_comparisons}, we illustrate how a simple context-based model is effective on CV point clouds but offers less gain on scientific particles. We start with a simple context model based on level marginals: $\hat{P}(O|\text{depth})$, and examine the coding gains achieved using an encoding distribution $\hat{P}(O|O_{\text{parent}}, \text{depth})$, which has a parent occupancy code as additional context. We observe that conditioning on parent occupancy code can significantly reduce coding costs for CV dataset MVUB at almost all levels, but it only mildly helps for scientific particle data Quijote. 

Beyond the gap to CV benchmarks, a practical question remains: do neural CV point cloud codecs improve over compressors already used in scientific computing? To answer that, we compare against two representative error-bounded scientific compressors, namely, SZ3~\cite{liang2022sz3} and LCP~\cite{zhang2025lcp}. SZ3 is a general-purpose lossy compressor that combines interpolation- and prediction-based coding. It is best suited for multi-dimensional field data, where values vary smoothly in a neighborhood of grid points. LCP, in contrast, is designed specifically for scientific particles. It re-orders and groups points by their spatial location to form small local blocks. This induces local correlation within each block and yields superior compression ratios to those of general-purpose scientific compressors. 

To make the comparison fair, we set the absolute error bound of both scientific compressors to half the position quantization step, which matches the worst-case error introduced by quantization in the neural codecs. The results in Table~\ref{tab:compression_bpp} show that SZ3 performs poorly. SZ3 predicts each value from its previously reconstructed neighbors and quantizes the prediction error into an integer code. Values whose prediction error exceeds the quantizer's range are marked as unpredictable and stored losslessly as raw floating-point numbers. On Quijote halos and particles, the spatially unordered coordinates produce extremely large prediction errors, so many points fall back to lossless storage. This results in a larger compressed size (compression ratio $<1$) than the post-quantization bpp. In contrast, LCP's compression ratio comes close to that of the best-performing CV compressors (G-PCC, RENO). This suggests that how points are ordered and grouped has a substantial impact on compression ratio, and we can look beyond the order imposed by the octree for scientific particle data.

\section{Perspectives: How can learning help scientific particle data compression in the future?}
\textbf{Neural position quantization} This work follows the common practice of uniformly quantizing position data to the point that the loss is negligible for certain downstream analyses. This is straightforward, but may not be optimal. Uniform quantization ignores that the importance of spatial regions can be non-uniform. A promising direction of machine learning is to replace uniform quantization with adaptive quantization that still preserves physically meaningful information. For example, topology-preserving compressors such as TopoSZp~\cite{agarwal2026toposzp} show that spatially adaptive quantization can preserve critical points in data while allowing aggressive bitrate savings. 

\textbf{Neural spatial partition and traversal} The octree imposes a uniform 8-way split at every node and a hard-coded traversal order. Predictive codecs in neural PCC consume occupancy codes from the most significant bits of the coordinates to the least, rather than following the native storage order. Both choices for partitioning and traversal order are practical for CV point clouds but mismatched to the geometries of scientific data. A promising direction is to learn both in a data-driven manner.

\textbf{Leveraging neural emulators as learned priors} Neural emulators for physical simulations can be repurposed as a learned prior for lossy compression. Emulators such as Jamieson \textit{et al.}~\cite{jamieson2023field} learn to map low-dimensional inputs (initial conditions and physical parameters) to full particle distributions. They can serve as physics-aware priors during predictive coding. This framework generalizes to more domains such as MD force fields~\cite{batzner20223} and plasma~\cite{gopakumar2024plasma}. 

\textbf{Bridging the CV and HPC communities} Progress on scientific particle compression will benefit from collaborations between two communities that work in parallel. The CV community brings sophisticated methods for 3D scene understanding and compression. The HPC community has deep domain expertise: which physical properties must be preserved, what distortion is tolerable, how data are generated on supercomputers, and how compression should fit in the data management pipeline. Concrete directions for collaboration include joint benchmarks that span both domains and co-designed codecs that are informed by the physics of the domain.
\appendix
\section{Scientific Dataset Curation and Preprocessing}\label{appendix:data}
Both octree-based codecs such as G-PCC and neural point cloud codecs operate on integer-valued 3D voxels, while scientific particle simulations store position data as floats. Applying existing point cloud codecs to scientific data requires us to bridge the gap from continuous coordinates to discrete grid positions. Across all datasets, we adopt a near-lossless preconditioning through uniform scalar quantization. A continuous coordinate $x$ is mapped to quantization index 
$x_i = \lfloor x / \Delta \rceil$. At de-quantization, quantization index is mapped back to continuous coordinate by $\hat{x}=\Delta\cdot x_i$. The quantization step size $\Delta$ is tailored to the requirement of the scientific domain, ensuring that the quantization error remains negligible for downstream analyses. Duplicate points arising from quantization are removed.

\paragraph{Cosmology $N$-body Simulations} The Quijote simulations~\cite{Quijote} is a large scale cosmology $N$-body simulation dataset designed to enable training of machine learning models for cosmology. It contains more than 40,000 simulations under various physical constraints and spatial resolutions. To cover different kinds of geometries, we derive three different particle datasets from the Quijote simulations: \textbf{halos}, \textbf{particles}, and \textbf{boxes}. All three variants are derived from 2000 simulations at redshift zero. Each simulation tracks the evolution of $512^3$ particles in a volume of $(1\, h^{-1}\,\mathrm{Gpc})^3$ with periodic boundary conditions. These 2000 simulations differ in 5 cosmological parameters, which were generated through Latin hypercube sampling. 

The \textbf{halos} dataset is a collection of subsampled dark matter halo position data. We read halo positions from the GADGET-format HDF5 files and convert to coordinates in $h^{-1}\,\mathrm{Mpc}$. From each snapshot, we uniformly subsample 100,000 halos at random using a per-realization random seed.

The \textbf{particles} dataset is a collection of subsampled dark matter position data. Similar to the \textbf{halos} dataset, we read particle positions from the GADGET-format HDF5 files and convert to coordinates in $h^{-1}\,\mathrm{Mpc}$. From each snapshot, we uniformly subsample 100,000 particles at random using a per-realization random seed.

The \textbf{boxes} dataset is a collection of particles inside a small cubic volume. From the same 2000 snapshots used to generate \textbf{halos} and \textbf{particles}, we extract particles within a small cubic block of volume $(125\, h^{-1}\,\mathrm{Mpc})^3$. We extract 8 randomly sampled blocks from each snapshot. This dataset is introduced to observe the compression performance of finer details of cosmology simulation, as opposed to \textbf{halos} and \textbf{particles}, which retain only larger scale structures.

For all three variants of Quijote data, we apply uniform scalar quantization to the positions as a near-lossless preconditioning prior to compression. The step size of $\Delta = 7.6\, h^{-1}\,\mathrm{Kpc}$ is chosen to be well below the gravitational softening length of the simulation. The softening length~\cite{springel2005cosmological} is the spatial scale below which gravitational forces are intentionally smoothed to prevent numeric explosion in close particle encounters. Therefore, no meaningful physical structure is resolved beneath this scale during simulation. Consequently, the information discarded by a quantization step even smaller than the gravitational softening length is effectively lossless for downstream analyses.

\paragraph{Molecular Dynamics} The YiiP equilibrium dataset~\cite{fan2019yiip} contains molecular dynamics (MD) trajectories of a membrane-protein system. It tracks the coordinate data of the entire system, including protein, membrane, water, and ions. Coordinate data are stored per atom for 111,815 atoms in 2 trajectories with 901 and 9001 frames, respectively. In order to acquire enough data to train neural compression models, we use the longer trajectory with 9001 frames.

The membrane-protein system was simulated with MD simulation code GROMACS~\cite{abraham2015gromacs}, and coordinate data are stored with 0.001 nm resolution. For this data, we apply quantization step of $\Delta = 0.001\,\text{nm}$, which is the same as the storage resolution. 

\paragraph{Plasma Physics} 
Sandberg \textit{et al.} \cite{sandberg2024synthesizing} proposed a dataset of charged particle beams in a plasma accelerator. The dataset tracks 1 million particles at 2 iterations, representing the initial and final snapshots. For each iteration, the particle position is recorded separately for 15 accelerator stages. We sequentially partition the 1 million positions at each stage into 100,000 point chunks, and create a dataset with 300 frames with 100,000 points each. We quantize the positions with $\Delta = 1\,\text{nm}$, which is small enough to result in less than 0.01\% point merging.

\begin{ack}
    This work was supported by the National Science Foundation (NSF) under grant number 2341055. This work was also supported by the U.S. Department of Energy, Office of Science, Advanced Scientific Computing Research (ASCR), under contract DE-AC02-06CH11357. This work was also supported by the U.S. National Science Foundation (OAC 2311875, OAC 2514036).
\end{ack}
\bibliographystyle{IEEEtran}
\bibliography{references}
\end{document}